\documentclass{mva_style}

\usepackage[utf8]{inputenc} 
\usepackage[T1]{fontenc}    
\usepackage{hyperref}       
\usepackage{url}            
\usepackage{booktabs}       
\usepackage{amsfonts}       
\usepackage{nicefrac}       
\usepackage{microtype}      
\usepackage{xcolor}         
\usepackage{graphicx}
\usepackage{algorithm}
\usepackage{algpseudocode}
\usepackage{amsmath}
\usepackage{bm}
\usepackage{booktabs}
\usepackage{multirow}
\usepackage{graphics}
\usepackage{caption}
\usepackage{xspace}
\usepackage{float}

\newcommand{\etal}{\emph{et al.}\xspace}

\finalcopy

\title{\LARGE \bf
    Modality Selection and Skill Segmentation via Cross-Modality Attention
}

\author{Jiawei Jiang$^1$, Kei Ota$^{2}$, Devesh K. Jha$^3$, and Asako Kanezaki$^1$}

\author{
  \bf{Jiawei Jiang} \\
  Science Tokyo \\
  \and
  \bf{Kei Ota}\\
  Mitsubishi Electric \\
  \and
  \bf{Devesh K. Jha}\\
  MERL \\
  \and
  \bf{Asako Kanezaki} \\
  Science Tokyo \\
}

\begin{document}
    \twocolumn[{%
    \renewcommand\twocolumn[1][]{#1}%
    \maketitle
    }]

    \section*{\centering Abstract}
\textit{
    Incorporating additional sensory modalities such as tactile and audio into foundational robotic models poses significant challenges due to the curse of dimensionality. This work addresses this issue through modality selection. We propose a cross-modality attention (CMA) mechanism to identify and selectively utilize the modalities that are most informative for action generation at each timestep. Furthermore, we extend the application of CMA to segment primitive skills from expert demonstrations and leverage this segmentation to train a hierarchical policy capable of solving long-horizon, contact-rich manipulation tasks.
}
    \section{Introduction}

Recent advances in language and vision foundation models \cite{gpt3, grattafiori2024llama3herdmodels, liu2023llava} have inspired the development of generalist foundation models for robotics \cite{octo_2023, open_x_embodiment_rt_x_2023, saycan2022arxiv}. These models are trained on large-scale datasets spanning diverse embodiments and tasks, demonstrating remarkable instruction-following capabilities. However, their manipulation performance---particularly in terms of precision and contact-rich interaction---remains limited.

A key limitation is the lack of manipulation-specific sensory inputs, such as tactile data, which are crucial for fine-grained manipulation. Simply increasing the number of sensory modalities by stacking inputs often leads to exponentially larger policy search spaces, making learning inefficient and unstable.

To address this, one line of work draws inspiration from Vision-Language Models (VLMs) \cite{liu2023llava, jones_beyond_2025}, projecting all modalities into the language domain to reduce search space dimensionality. While this improves generalization in simple scenarios like pick-and-place, it often sacrifices modality-specific granularity, making it unsuitable for fine-grained, contact-rich tasks.

In contrast, we take inspiration from how humans manage multi-modal sensory inputs. Despite receiving a vast array of sensory signals, humans dynamically focus on the most relevant modalities depending on the task at hand. This selective attention is mirrored in the self-attention mechanisms of transformers, which can adaptively filter and prioritize information across modalities.

In this paper, we explore this idea by applying a transformer-based multi-modal policy architecture to furniture assembly tasks~\cite{heo2023furniturebench}, which demand high accuracy and nuanced manipulation. Our goal is to improve the efficiency and accuracy of training multi-modal imitation learning policies for long-horizon, contact-rich tasks. We summarize our contributions as follows:
\begin{itemize}
    \item{\textbf{Cross-Modality Attention (CMA)}: We propose a transformer-based architecture with a conditional U-Net backbone to learn CMA through imitation learning.}
    \item{\textbf{Unsupervised Primitive Segmentation}: We demonstrate that CMA enables unsupervised segmentation of expert trajectories into primitive actions.}
    \item{\textbf{Improved Sample Efficiency}: We show that training on segmented primitive actions leads to better accuracy and sample efficiency compared to end-to-end trajectory learning, due to more consistent input distributions within each primitive.}
\end{itemize}

    \section{Related Work}

\noindent \textbf{Diffusion Policy.}
Diffusion models \cite{ho_denoising_2020, song2021scorebased} have emerged as a powerful class of generative models, revolutionizing data synthesis in domains such as image, audio, and more recently, robotics. Chi \etal \cite{chi2024diffusionpolicy} pioneered the application of conditional denoising diffusion to the robot action space, addressing key challenges in imitation learning---namely, multi-modal action distributions, temporal correlation, and the high precision demands of robotic tasks. This framework has since inspired a surge of research exploring variations of diffusion-based policies in robotics \cite{ze_3d_2024, ankile_juicer_2024}. Further improvements have been proposed through goal-conditioned or reward-guided diffusion models \cite{reuss2023goal, ren_diffusion_2024}, aiming to inject task-specific guidance during action generation.

\noindent \textbf{Multi-modal training for Robotics.}
Recent work on generalist robotic models \cite{octo_2023, open_x_embodiment_rt_x_2023, saycan2022arxiv} has demonstrated the potential of using vision and language inputs to train policies capable of performing a wide range of manipulation tasks. These models, trained on large-scale datasets, exhibit impressive generalization. However, contact-rich manipulation remains a challenge due to its complexity and the limited information available from vision alone.

To address this, a growing body of work has focused on multi-sensory learning. The ObjectFolder series \cite{gao_objectfolder_2021, gao_objectfolder_2022, gao_objectfolder_2023} introduced simulated environments incorporating tactile and audio modalities. Li \etal~\cite{li_see_2022} showed that incorporating such modalities improves performance in tasks like dense packing and water pouring. Similarly, Feng \etal~\cite{feng2024play} introduced a method that uses state tokens to dynamically adjust modality importance during task execution. Jones \etal~\cite{jones_beyond_2025} proposed a framework that enables multi-modal prompting by fine-tuning generalist models with tactile and audio inputs, extending their utility beyond vision and language.

\noindent \textbf{Skill Composition for long horizon tasks.}
Long-horizon manipulation tasks are often tackled by composing primitive actions. In traditional approaches, model-based frameworks such as behavior trees \cite{colledanchise_behavior_2018} and task-and-motion planning \cite{garrett_integrated_2021, curtis_2022_motion_planning, driess_learning_2021} explicitly define this structure. In contrast, learning-based methods \cite{luo_multistage_2024, mao_dexskills_2024} seek to learn hierarchical policies composed of a skill selector and a library of primitive skills. However, these approaches typically rely on human-labeled segmentation of demonstration trajectories, which is not only labor-intensive but also susceptible to ambiguity and inconsistency.

\noindent \textbf{Furniture Assembly.}
Furniture assembly has become a widely studied benchmark for long-horizon, contact-rich manipulation due to its complexity and structured sequence of operations. Heo \etal~\cite{heo2023furniturebench} introduced a standardized simulation environment and a suite of tasks to evaluate robotic performance in this domain. Building on this, Ankile \etal \cite{ankile_juicer_2024} identified bottleneck states in expert demonstrations to generate synthetic trajectories, enriching training data diversity. Other approaches include the use of hierarchical reinforcement learning, such as in Lin \etal~\cite{lin_generalize_2024}, which combines offline-trained skill transition models with skill-conditioned policies. Additionally, Ankile \etal~\cite{ankile2024imitationrefinementresidual} proposed a residual policy learning framework \cite{silver2019residualpolicylearning} to refine diffusion model outputs with reinforcement learning corrections.

Despite these advancements, furniture assembly remains a challenging domain, underscoring the need for improved modeling of temporal structure, modality fusion, and generalization in contact-rich tasks.
    \section{Method}
To investigate our hypothesis that CMA can be used to segment primitive actions in an unsupervised fashion after sufficient training, we design a model that integrates CMA with a 1D-conditional U-Net diffusion policy. The goal is to leverage expert demonstrations via imitation learning and to train CMA to extract task-relevant features that can enhance policy performance in downstream contact-rich manipulation tasks.

\begin{figure*}
    \centering
    \includegraphics[width=0.8\linewidth]{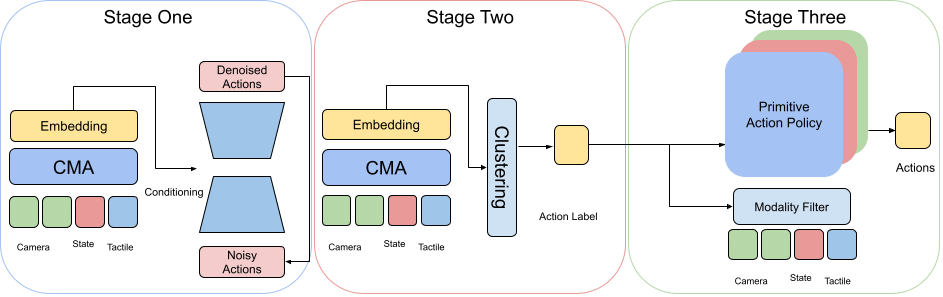}
    \caption{Three stages design. First stage: train CMA using imitation learning. Second stage: Use CMA to cluster and segment primitive actions. Thrid stage: Train a hierarchical policy that select and execute individual primitive actions that only select useful modalities as input. }
    \label{fig:architecture}
\end{figure*}

Our architecture consists of a 1D-conditional U-Net diffusion model as the policy backbone. We process state inputs using a fully connected layer and image inputs using a frozen R3M encoder~\cite{nair_r3m_2022} with a trainable projection layer. The resulting embeddings from each modality and timestep are stacked into a tensor of shape $(T \cdot N \times B \times D$, where T is the number of timesteps, N is the number of modalities, B is the batch size and D is the embedding dimension, which is passed through the CMA module. The output of the CMA serves as the conditional embedding for the diffusion model, which then generates action chunks for execution (see Fig.~\ref{fig:architecture}).

\subsection{Multi-modal Diffusion Policy}
We adopt a 1D-conditional U-Net diffusion model \cite{chi2024diffusionpolicy} as the backbone of our policy due to its demonstrated training stability and ability to model temporally correlated, multi-modal action distributions. The diffusion model outputs actions in discrete chunks, which improves the smoothness and consistency of rollouts. Following the architecture from Chi \etal~\cite{chi2024diffusionpolicy}, we use 256, 512, and 1024 as the downsampling dimensions of the U-Net.
\subsection{Cross-Modality Attention}
To fuse information from multiple modalities and timesteps, we integrate CMA into our policy. Inspired by the self-attention mechanisms in transformers, CMA enables the model to dynamically weight and integrate inputs from different modalities in a shared embedding space. This mechanism supports contextual fusion based on task relevance.

Prior work \cite{li_see_2022} has shown that cross-modal attention outperforms simple concatenation in multi-modal robotic tasks such as dense packing and water pouring. We follow this design by incorporating CMA into our conditioning pipeline. To further capture temporal context, we stack observations from two consecutive timesteps, allowing CMA to attend across both modalities and time.

For the CMA module, we use 8 attention heads and 2 transformer layers, which we found sufficient to model relevant cross-modal interactions in our task setup.

\subsection{Encoder}

\noindent \textbf{Image Encoder.}
We use the R3M encoder \cite{nair_r3m_2022}, based on ResNet-18, to process visual observations. The R3M weights are frozen, and we add a trainable fully connected layer to project the visual embeddings into a 128-dimensional latent space.

\noindent \textbf{State Encoder.}
State observations are processed using a fully connected layer that projects them into the same 128-dimensional space as the image embeddings. This design allows both modalities to be aligned in a shared embedding space \cite{liu2023llava}, facilitating effective fusion.

\section{Experiments}
\begin{figure*}[t]
    \centering
    \includegraphics[width=0.7\linewidth]{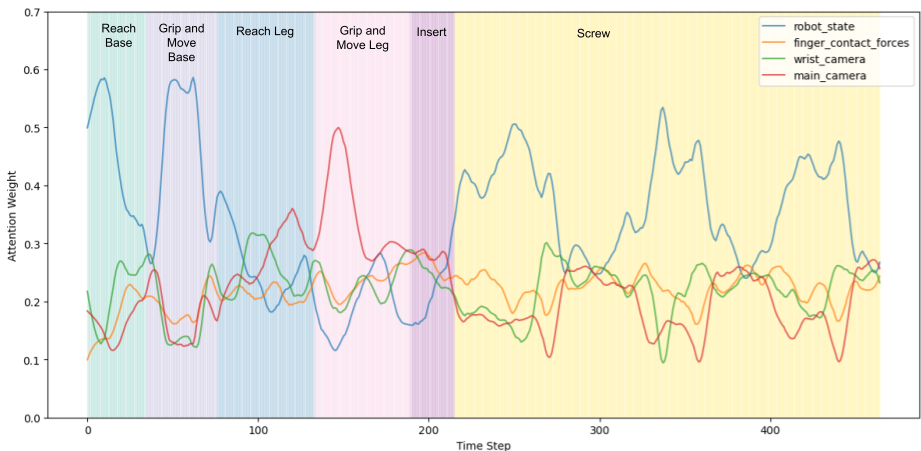}
    \caption{Attention weights obtained through training a multimodal diffusion model.} 
    \label{fig:attention_weights}
\end{figure*}
\subsection{Settings}

In this section, we investigate our hypothesis that CMA can be used to segment primitive actions in an unsupervised fashion after sufficient training by answering the following questions:
\begin{enumerate}

\item Does CMA learn characteristics of different primitive actions?
\item Does training primitive actions improve efficiency?
\item Can we use unsupervised methods to cluster primitive actions from raw data?
\end{enumerate}

\noindent \textbf{Environment.}
We evaluate the proposed method on the \emph{FurnitureSim}~\cite{heo2023furniturebench} environment, specifically the one leg assembly task where the robot is tasked to assemble a single leg onto the square table base. We choose \emph{FurnitureSim} as our simulation environment for its complex and long-horizon nature. We believe multi-modal data are particularly useful for solving these types of tasks.

\noindent \textbf{Expert Data.}
We obtain 50 trajectories of expert data from a state based policy trained in Ren \etal~\cite{ren_diffusion_2024} which achieved robust accuracy and high success rate (96\%). We manually segment the trajectories into 6 primitive actions: \emph{Reach Base}, \emph{Grip and Move Base}, \emph{Reach Leg}, \emph{Grip and Move Leg}, \emph{Insert}, and \emph{Screw}, similar to what's defined in Lin \etal~\cite{lin_generalize_2024}.
\begin{figure}[t]
    \centering
    \includegraphics[width=\linewidth]{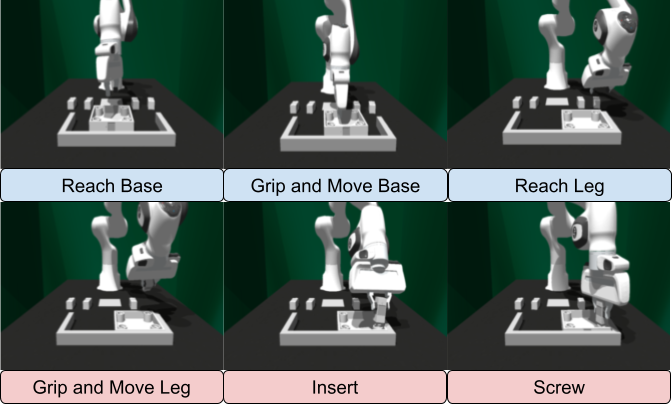}
    \caption{Primitive Actions.}
    \label{fig:primitive_actions}
\end{figure}

\noindent \textbf{Baseline Model.}
We train a multi-modal diffusion policy that takes as input third-person camera, grip camera, proprioceptive sensory data and tactile data to output end-effector pose using the expert data. We use the same batch size (32) as Ankile \etal~\cite{ankile2024imitationrefinementresidual} and limit the total training steps to 1,200,000 steps.
\subsection{Attention characteristics of different primitive actions}
We use the following experiment to examine if the CMA has learned the characteristics of the primitive actions previously defined. After the baseline model is trained, we use the transformer in it to process a whole trajectory from the expert data and record the attention weights of each individual modality. 
Formally, the self-attention is formulated as,
$$
\begin{aligned}
\text{Attention}(Q, K, V) &= \underbrace{\text{softmax} \left( \frac{QK^\top}{\sqrt{d_k}} \right)}_{\text{Attention Weights}} \cdot V \\
\end{aligned}
$$
We customize the self-attention block in the CMA to return both the attention weights and the final output. We take the attention weights of the last layer, averaged across all heads, for our experiment as it is most directly connected to the final decision making. 
The pattern of the attention weights show clear differences across various primitive actions, as shown in Figure \ref{fig:attention_weights}.

\subsection{Training Whole Trajectory v.s. Individual Primitives}
\begin{figure}[t]
    \centering
    \includegraphics[width=\linewidth]{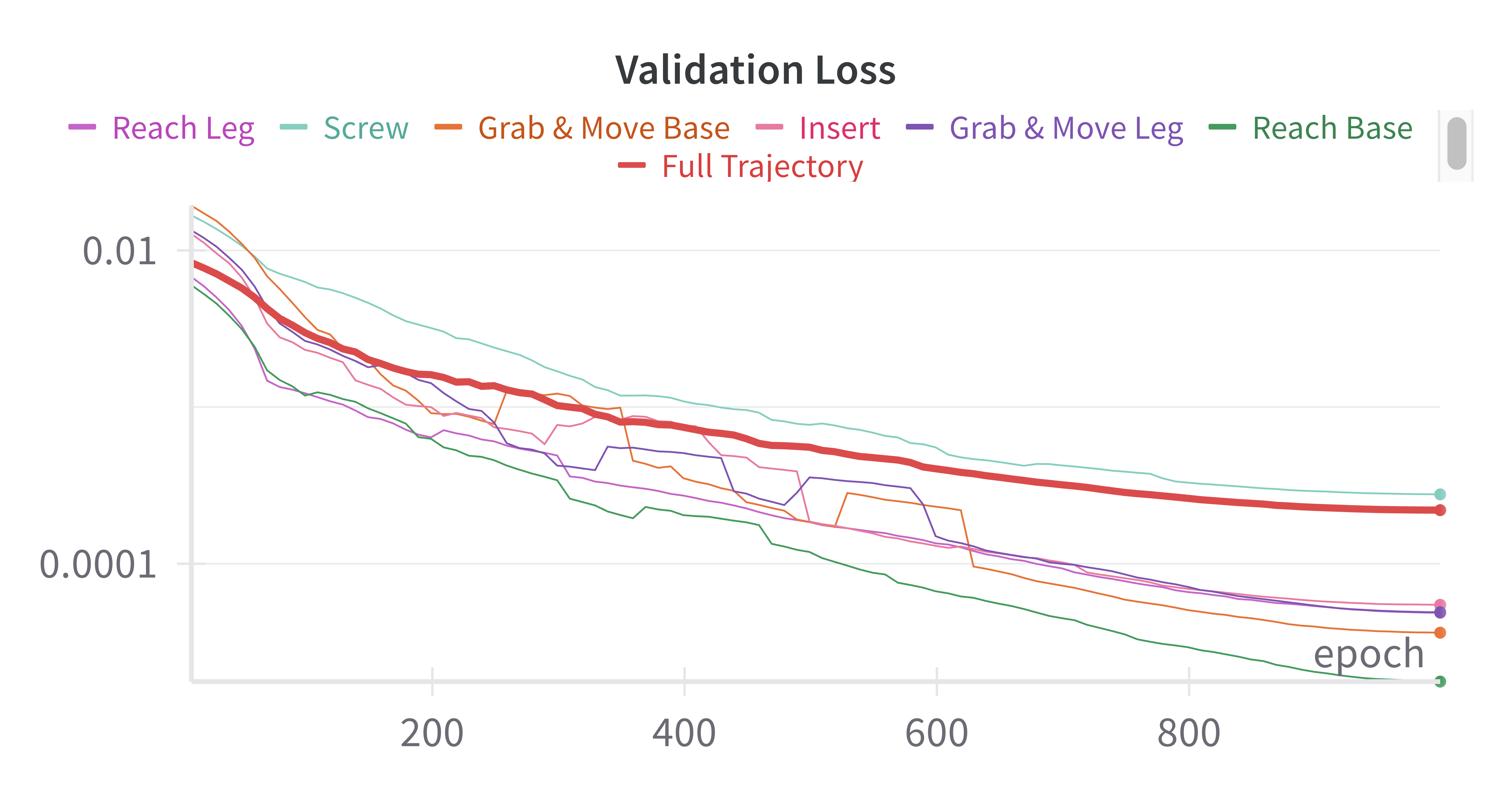}
    \caption{Validation Loss.}
    \label{fig:results_training_curves}
\end{figure}
In this section, we show the results of training a single policy using the entire trajectory v.s. training 6 different policies for each individual primitive action in Figure \ref{fig:results_training_curves}. 
We train each individual policy for 200,000 steps to ensure the total training steps match those of the baseline model. 
The validation loss graph shows that training policies for primitive actions drastically improve sample efficiency and accuracy except the screwing action, which we will explain in Section \ref{sec:tsne}.

\subsection{Clustering action segments}
\label{sec:tsne}
\begin{figure}[t]
    \centering
    \includegraphics[width=\linewidth]
    {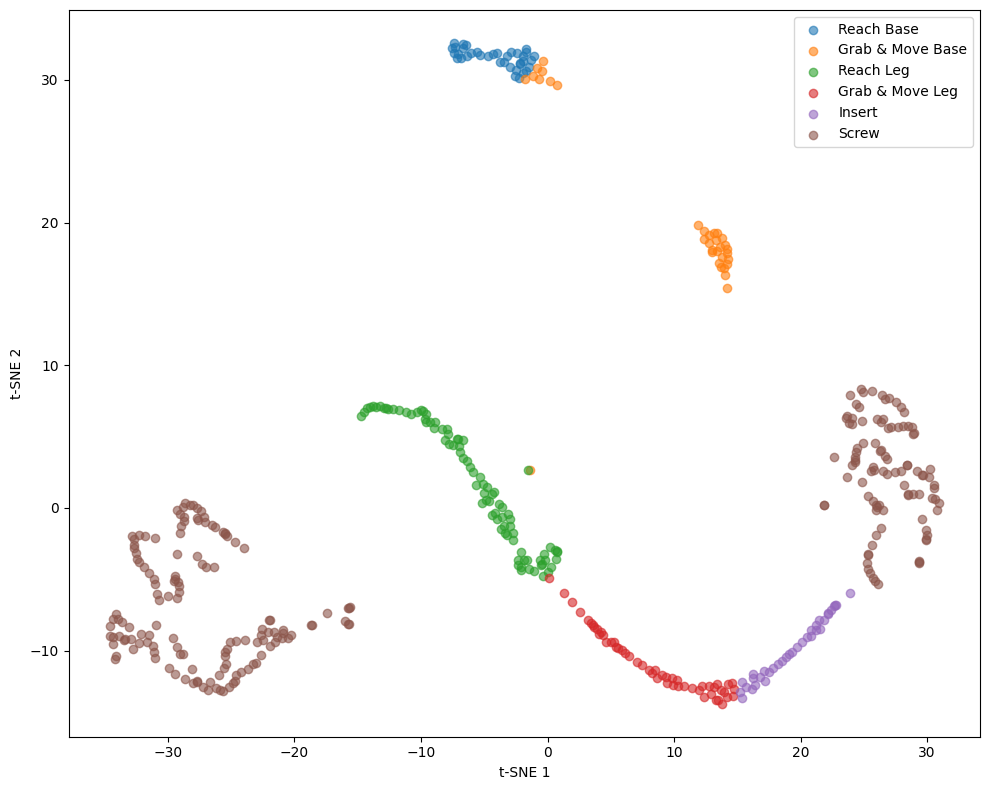}
    \caption{T-SNE graph of primitive actions}
    \label{fig:tsne}
\end{figure}

Here we visualize the embedding space of the primitive actions using T-SNE for an entire trajectory. We can see that the embeddings have been separated into clusters. However, due to the ambiguity in human labels the clustering is not perfect. For example, some actions from Grab \& Move are grouped into the Reach Base action, and the Screw action has been grouped into two distinct clusters, representing the screwing and reposition of the gripper. These observations suggest that the flaw of relying on human labels for action segmentation, which is likely the cause of underperformance of the Screwing policy specifically shown in Figure \ref{fig:results_training_curves}.
It may be more effective to segment the trajectory using unsupervised methods.


    \section{Conclusion}
Our results confirmed our hypothesis and answered three crucial questions for future research: 
\begin{enumerate}
    \item Cross Modality Attention learns characteristics of different primitive actions through the attention weights.
    \item Training individual primitive actions drastically improve training efficiency, because of the massively different input distribution among individual primitive actions.
    \item The embedding visualization showed us how the CMA perceive actions differently from human, and confirmed that it is better to segment actions in an unsupervised manner.
\end{enumerate}

In future research, we plan to use unsupervised methods to segment the primitive actions and train an hierarchical policy to select and execute the individual primitive action policies based on our findings. Furthermore, we aim to limit input modalities to each individual primitive policy based on the attention weights to further improve the sample efficiency. 
    \bibliographystyle{IEEEtran}
    \bibliography{reference}

\end{document}